\documentclass[journal]{IEEEtran}
\usepackage{amsmath,amsfonts,amssymb}
\usepackage{algorithmic}
\usepackage{algorithm}
\usepackage{array}
\usepackage[caption=false,font=normalsize,labelfont=sf,textfont=sf]{subfig}
\usepackage{textcomp}
\usepackage{stfloats}
\usepackage{url}
\usepackage{verbatim}
\usepackage{graphicx}
\usepackage{cite}
\usepackage{multirow}
\usepackage{booktabs}
\usepackage{xcolor}
\usepackage{bm}
\usepackage{hyperref}
\begin{document}
\title{Dual-Integrated Low-Latency Single-Lens Infrared Computational Imaging for Object Detection}

\author{Xuquan Wang, Guishuo Yang, Dapeng Yan, Yujie Xing, Xuanyu Qian, Kai Zhang, Xiong Dun, and Jiande Sun%
\thanks{Xuquan Wang and Guishuo Yang contributed equally to this work. Corresponding authors: Xiong Dun and Jiande Sun.}%
\thanks{Xuquan Wang, Guishuo Yang, Dapeng Yan, Yujie Xing, Xuanyu Qian, and Xiong Dun are with the MOE Key Laboratory of Advanced Micro-Structured Materials, the Institute of Precision Optical Engineering, School of Physics Science and Engineering, Tongji University, and the Shanghai Frontiers Science Center of Digital Optics, Shanghai 200092, China.}%
\thanks{Kai Zhang and Jiande Sun are with the School of Computer Science and Artificial Intelligence, Shandong Normal University, Jinan 250358, China, and the Shandong Engineering Research Center for Multimodal Computing and Intelligent Decision Making, Shandong Normal University, Jinan 250358, China.}%
\thanks{E-mail: dunx@tongji.edu.cn; jiandesun@hotmail.com.}}

\markboth{Journal of \LaTeX\ Class Files,~Vol.~14, No.~8, August~2021}{Wang \MakeLowercase{\textit{et al.}}: Dual-Integrated Infrared Computational Imaging}

\maketitle
\begin{abstract}
Computational imaging enables high-quality infrared imaging with compact optical architectures by reducing the complexity of optical path design through deep learning. This deep-learning-based process typically involves both reconstruction and detection. Although the use of deep learning improves detection performance, it also increases the computational cost and inference latency of the imaging system. Most existing methods focus on reducing the complexity of the deep-learning pipeline, while ignoring the guidance provided by the physical priors of the optical path, and therefore remain constrained by the trade-off between accuracy and latency. To address this limitation, we present a Physics-aware Dual-Integrated Network (PDI-Net), which integrates physical priors from the optical path with deep learning while further integrating infrared image reconstruction and object detection, thereby enabling accurate object detection with low latency. Specifically, a U-Net is supervised by ground truth, and the semi-U-Net is designed as a feature-sharing layer that is directly connected to the YOLO backbone, enabling accurate object detection with low latency. A physics-aware large-small bridge (PALS-Bridge) is introduced to bridge the gap between fidelity-oriented features in the semi-U-Net and detection-oriented semantic representations in YOLO. Motivated by the observation that field-dependent point spread functions (PSFs) determine the spatially varying degradation intensity in single-lens imaging, the bridge uses PSF priors to adaptively modulate multiscale convolutional branches for targets of different sizes. Furthermore, a physics-informed optical degradation simulation pipeline is developed to support training and validation. Finally, PDI-Net is deployed on a single-lens infrared camera, achieving an approximately 50\% reduction in system weight compared with traditional multi-lens designs. Compared with the \textit{Rec+Det with pruning} strategy in the low-SNR setting, the proposed method reduces inference time by 84.06\% while improving mAP@0.5:0.95 by 5.07\%. This work paves the way for compact, high-performance computational infrared imaging and real-time video-level target detection in resource-constrained environments.
\end{abstract}

\begin{IEEEkeywords}
computational imaging, object detection, image reconstruction, joint optimization, edge AI, single-lens infrared imaging
\end{IEEEkeywords}

\section{Introduction}
	Infrared imaging plays a vital role in industrial applications and the Internet of Things (IoT), including smart cities, precision agriculture, autonomous driving, and security monitoring, owing to its ability to operate in low-light and harsh environments without active illumination \cite{Cai2024TIP, zhou2021assessment, jiang2022object}. The miniaturization and integration of infrared cameras are becoming increasingly crucial to meet the requirements of perception-layer platforms such as lightweight unmanned aerial vehicles (UAVs) and wearable devices \cite{jiang2022object, Miniaturization}. However, traditional optical systems struggle to balance high performance and lightweight design, as they require multiple lenses for geometric aberration correction \cite{Laskin2021}. Directly reducing the number of lenses can significantly degrade image quality. Computational imaging is an innovative optical paradigm that integrates optical acquisition with computational algorithms through joint design \cite{Heide2013, bhandari2022computational, liu2023research}. By embedding computation into the imaging process, it significantly relaxes the constraints of traditional optical systems \cite{sinha2017lensless, suo2023computational}. As a result, high-quality imaging, comparable to that of complex optical setups, can be achieved using simpler and more compact configurations. With phase differences corrected by backend algorithms, a variety of single-lens imaging strategies have emerged, including deep Fresnel lenses, diffractive optical elements, and metasurfaces \cite{zhang2023end, Bian2022, Zuo2022, dun2020learned, li2023lightridge}. In addition, lensless systems have also emerged as a new paradigm \cite{Lensless, asif2016flatcam, antipa2017diffusercam}. Although these approaches simplify system architecture, single-lens computational imaging inevitably introduces additional costs. Specifically, the use of specialized optical designs for information encoding necessitates corresponding image reconstruction algorithms to restore high-quality images \cite{he2010single, yuanxin}. This leads to increased latency, as well as higher computational and power demands, which limit the applicability of such systems in high-speed UAV scenarios under resource-constrained conditions.
	
	Model compression for the image reconstruction process is a conventional approach to accelerating computational imaging at the edge \cite{deng2020model}. Lightweight compression techniques, such as pruning \cite{han2015deep}, quantization \cite{Quantization_Lee, jacob2018quantization}, distillation \cite{hinton2015distilling}, and neural architecture search \cite{pham2018efficient}, have demonstrated promising results in simplified and compact computational imaging applications. In our early research, we achieved significant optimization of reconstruction time through model pruning, enabling video-level imaging at 25 frames per second (FPS) \cite{xing2025real, Xing_Physics-informed, MWR-Net}. Considering that implementing image reconstruction algorithms on edge AI chips is a complex, multi-faceted issue involving factors such as operator configuration, chip architecture, hardware design, and memory access bottlenecks, we further proposed an edge-accelerated reconstruction strategy based on end-to-end sensitivity analysis for single-lens computational imaging \cite{Wang2025}. Compared with uniform pruning on the GPU, edge sensitivity-guided reconstruction algorithms achieve simultaneous improvements in both reconstruction quality and inference speed. Nevertheless, these efforts mainly focus on backend optimization of the reconstruction pipeline. For resource-constrained conditions, such optimization can alleviate the computational burden, but it does not explicitly exploit the physical priors embedded in the imaging process, such as field-dependent degradation properties introduced by single-lens design. As a result, further improvements in latency and downstream detection accuracy remain fundamentally limited.
	
\begin{figure*}[!t]
		\centering
		\includegraphics[width=\textwidth]{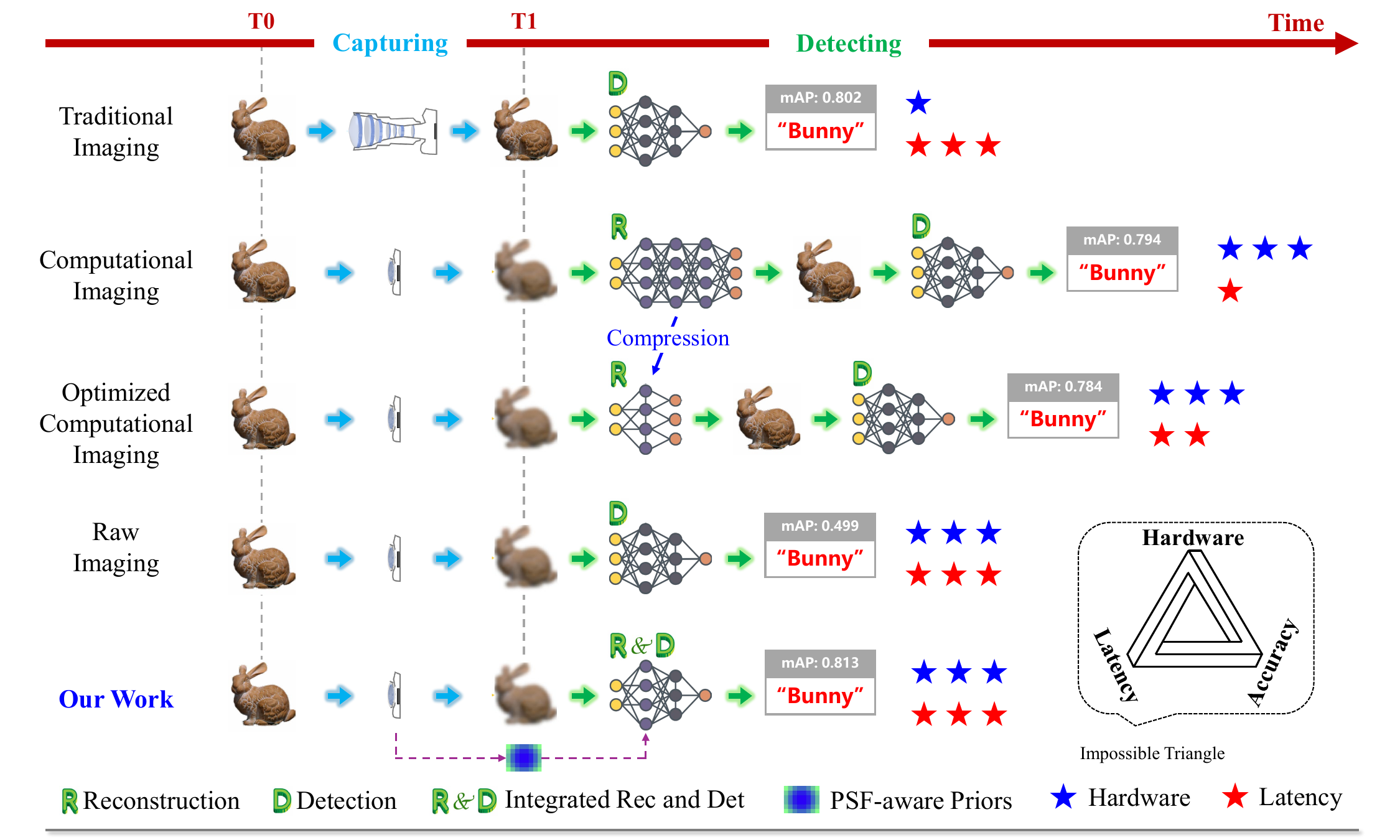}
		\caption{A comparative illustration of infrared object detection methodologies employing distinct imaging strategies.}
		\label{fig:strategy}
	\end{figure*}
	
	In recent years, researchers in machine vision have realized that object detection does not necessarily require reconstructing the entire image \cite{sun2022rethinking, low2004distinctive, dalal2005histograms, girshick2014rich, girshick2015fast}. Instead, extracting essential features, such as edges, textures and shapes, is often sufficient for subsequent processing \cite{liu2017image, GAN, yang2023cross}. From an information-theoretic perspective, information acquisition in imaging is completed once the detector finishes exposure and photoelectric conversion. The reconstruction process does not generate new informative content; rather, it reorganizes the encoded measurements into formats that are more interpretable to humans or compatible with conventional downstream algorithms. This understanding has been preliminarily validated in traditional image restoration tasks, such as denoising \cite{dong2018denoising, jo2021rethinking, liang2024image}, dehazing \cite{cai2016dehazenet, hu2024beyond} and rain removal \cite{wang2021rain}, and has further inspired the development of limited information distribution theory in compressive spectral imaging \cite{yang2024compressive}. Inspired by the analysis above, we assume that the raw encoded output of a lightweight computational imaging camera can be directly exploited for object detection through integrated reconstruction and detection. Consequently, the additional latency introduced by full image reconstruction could, in principle, be avoided.
	
	However, directly applying conventional integrated reconstruction and detection algorithms often yields unsatisfactory results. Unlike traditional image restoration, where degradations mainly arise from natural phenomena such as fog, rain, or motion blur \cite{yuan2024bi, li2017aod, li2023detection}, and are commonly addressed using statistical priors or data-driven techniques \cite{jin2022darkvisionnet, zhang2023learning, cui2021multitask, hnewa2021multiscale, marathe2022restorex}, computational imaging is fundamentally enabled by the coupling between frontend optical design and backend reconstruction algorithms \cite{fu2024blind}. The degradation mechanisms are more complex, encompassing optical aberrations, sensor noise, and artifacts introduced by compressive sampling. As a result, the reconstruction process in computational imaging is inherently dependent on the physical priors of the imaging system. Therefore, simply integrating reconstruction and detection is insufficient for lightweight single-lens infrared imaging; physical priors must also be explicitly incorporated into the learning pipeline. Meanwhile, a substantial feature-level gap remains between reconstruction tasks, which emphasize fidelity-oriented features, and detection tasks, which rely on discriminative semantic representations. This discrepancy complicates direct feature sharing and undermines model robustness \cite{DeepDenoising}.
		
	\begin{figure*}[!b]
		\centering
		\includegraphics[width=\textwidth]{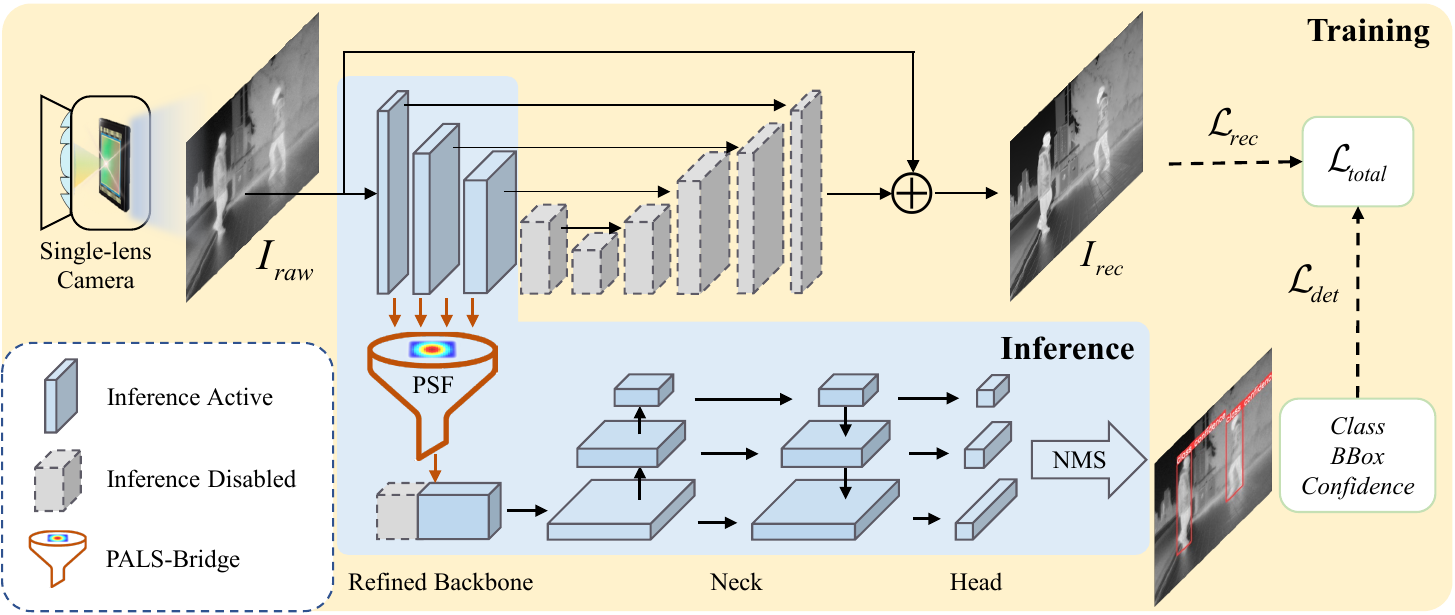}
		\caption{Overview of the proposed PDI-Net for low-latency single-lens infrared computational imaging.}
		\label{fig:framework}
	\end{figure*}
	
	To further illustrate the evolution and trade-offs of these strategies, Fig.~\ref{fig:strategy} provides a multi-dimensional comparison, including traditional multi-lens imaging for object detection \cite{Laskin2021}, single-lens computational imaging with separate reconstruction and detection \cite{zhang2023end}, single-lens computational imaging with compressed reconstruction and subsequent detection \cite{Wang2025}, single-lens imaging with direct detection on raw degraded inputs, and the proposed Physics-aware Dual-Integrated Network (PDI-Net). The horizontal axis depicts the total inference timeline, encompassing both capturing and detection phases. Specifically, each strategy is evaluated across three key metrics: hardware complexity, inference latency, and detection accuracy. These three metrics constitute a practical Impossible Triangle, as illustrated in Fig.~\ref{fig:strategy}, meaning that none of them can be simultaneously optimized to their fullest extent. Notably, the first four strategies do not explicitly incorporate physical priors and therefore mainly seek a balance between latency and accuracy within a conventional trade-off boundary. By contrast, the proposed PDI-Net integrates physical priors into the learning-based algorithm, which helps break the conventional boundary and achieve a more favorable balance under resource-constrained conditions.
	
	To address the above challenges, we develop a Physics-aware Dual-Integrated Network, named PDI-Net, for low-latency single-lens infrared computational imaging. The proposed framework achieves dual integration from both the task and physical perspectives: it not only integrates infrared image reconstruction with object detection, but also integrates physical priors of the imaging process into the learning-based algorithm. Specifically, a U-Net is introduced to learn the mapping from degraded measurements to ground truth images, while a partial encoder, referred to as the semi-U-Net, serves as a shared feature extractor and is directly connected to the YOLO backbone. To bridge the representational gap between fidelity-oriented features in the semi-U-Net and detection-oriented semantic representations in YOLO, we further introduce a physics-aware large-small bridge (PALS-Bridge) together with a joint optimization strategy. Motivated by the observation that field-dependent point spread functions (PSFs) determine the spatially varying degradation in single-lens imaging, PALS-Bridge integrates PSF priors to adaptively modulate multiscale convolutional branches for targets of different sizes. Furthermore, a physics-informed simulation pipeline is developed to synthesize optical degradation datasets, thereby embedding physically meaningful priors into the training process. The proposed PDI-Net is deployed on an RK3588 edge AI chip integrated with a single-lens infrared computational imaging camera, resulting in an approximately 50\% reduction in system weight compared with traditional multi-lens designs, and is further validated through flight experiments on a small UAV. Compared with the \textit{Rec+Det with pruning} strategy in the low-SNR setting, PDI-Net reduces inference time by 84.06\% while improving mAP@0.5:0.95 by 5.07\%. These results demonstrate that the proposed approach enables compact, high-performance infrared computational imaging and real-time video-level target detection on resource-constrained platforms.

\section{Proposed method}
In this section, we provide a detailed introduction to the PDI-Net for low-latency single-lens infrared computational imaging, along with its joint optimization strategy and dataset generation paradigm.
\subsection{Overall framework}
	
	Fig.~\ref{fig:framework} illustrates the overall architecture of the proposed PDI-Net for low-latency single-lens infrared computational imaging, which integrates reconstruction and detection while further integrating physical priors with the learning-based algorithm. The framework comprises a U-Net-based reconstruction module, a YOLO-based detection module, and a physics-aware connector termed PALS-Bridge. During training, as indicated by the yellow background in Fig.~\ref{fig:framework}, the input is a degraded infrared image ${I_{raw}}$ from the synthetic dataset. The U-Net is used to reconstruct a clear image ${I_{rec}}$ under the supervision of the reconstruction loss ${\cal L}_{rec}$ against the ground truth. Considering the high computational cost of full reconstruction, the early encoder layers, referred to as the semi-U-Net, are redirected to the fine-grained detection backbone. However, the substantial gap between low-level, pixel-aligned textures and high-level semantic features makes direct feature sharing unstable~\cite{DeepDenoising}. To address this, the proposed PALS-Bridge is introduced to align these heterogeneous feature representations. The detection head outputs class predictions, bounding boxes, and confidence scores to compute the detection loss ${\cal L}_{det}$. The total loss ${\cal L}_{total}$ is formulated as a weighted sum of ${\cal L}_{rec}$ and ${\cal L}_{det}$. During inference, as highlighted in blue in Fig.~\ref{fig:framework}, the input can be either a degraded image from the dataset or a raw measurement captured by the single-lens camera. Only the semi-U-Net, PALS-Bridge, and detection components are executed, thereby bypassing full reconstruction and significantly reducing inference latency.
	
	\subsection{Reconstruction module}
	Infrared images typically suffer from low signal-to-noise ratio (SNR), especially in single-lens computational imaging systems. Compared with traditional computational imaging reconstruction networks, the primary role of our reconstruction module has shifted from producing clear images to efficiently extracting shared features for joint tasks. Owing to its symmetric encoder-decoder architecture and favorable trade-off among reconstruction quality, inference speed, and suitability for edge deployment~\cite{dun2020learned,xing2025real,Wang2025,MWR-Net,Xing_Physics-informed}, we adopt an enhanced U-Net as the reconstruction module. Its multiscale encoder-decoder structure facilitates feature extraction, while the skip connections help preserve fine-grained spatial information, such as edges, textures, and object contours, that are critical for accurate downstream detection. The detailed U-Net architecture is provided in Supplementary Fig. S1.
	
	During the training phase, the reconstruction module is jointly optimized as an integral component of the overall framework. Its optimization is primarily guided by a loss function constructed from pairs of degraded infrared images ${I_{raw}}$ and their corresponding clear ground truth ${I_{gt}}$. The specific formulation of the loss function ${\cal L}_{rec}$ is defined as
	\begin{equation}
		\mathcal{L}_{rec} = \frac{1}{N}\sum\limits_{i = 1}^N \left\| I_{rec}^{(i)} - I_{gt}^{(i)} \right\|_2^2,
		\label{eq:rec-loss}
	\end{equation}
	where $I_{rec}^{(i)}$ and $I_{gt}^{(i)}$ are the reconstructed image and ground truth image for the $i$-th sample, respectively, $\left\| \cdot \right\|_2^2$ denotes the $L_2$ norm, and $N$ is the total number of images in a batch.
	
	\subsection{Detection module}
	The design of the object detection module follows principles similar to those of the reconstruction module, requiring a balance between detection accuracy, computational efficiency, and edge deployment compatibility. As a representative architecture for real-time object recognition, the YOLO series has demonstrated remarkable effectiveness across various applications and has undergone extensive structural optimization for edge efficiency~\cite{redmon2016you,ge2021yolox,wang2024yolov10}. In this work, we adopt a streamlined and optimized version of the classical YOLOv5 architecture as the core detection module. Furthermore, to validate the generality of the proposed integration framework across different YOLO variants, we also evaluate the performance of an extended YOLOv8-based implementation.
	
	In the proposed PDI-Net, the shared features extracted from the reconstruction module are directly transmitted to the refined detection backbone. To facilitate the transfer and fusion of heterogeneous features, the backbone structure is optimized to maintain consistency in channel dimensions and spatial resolutions. The detailed connection strategies are discussed in the ablation study in Section 3.4. The detection module is supervised by a compound loss function ${\cal L}_{det}$, defined as
	\begin{equation}
		\mathcal{L}_{det} = hyp_{box} \mathcal{L}_{box} + hyp_{obj} \mathcal{L}_{obj} + hyp_{cls} \mathcal{L}_{cls},
		\label{eq:det-loss}
	\end{equation}
	where the hyperparameters $hyp_{\textit{box}}$, $hyp_{\textit{obj}}$, and $hyp_{\textit{cls}}$ serve as weighting coefficients for the bounding box regression, objectness confidence, and classification loss terms, respectively. ${\cal L}_{box}$, ${\cal L}_{obj}$, and ${\cal L}_{cls}$ denote the corresponding detection losses used in the YOLO-based detector. This integrated detection pipeline leverages shared feature representations and task-specific loss optimization, ensuring efficient and accurate target detection even under the highly challenging conditions inherent to single-lens infrared computational imaging.
	
	\subsection{Physics-Aware Large-Small Bridge}
\begin{figure*}[!t]
		\centering
		\includegraphics[width=\textwidth]{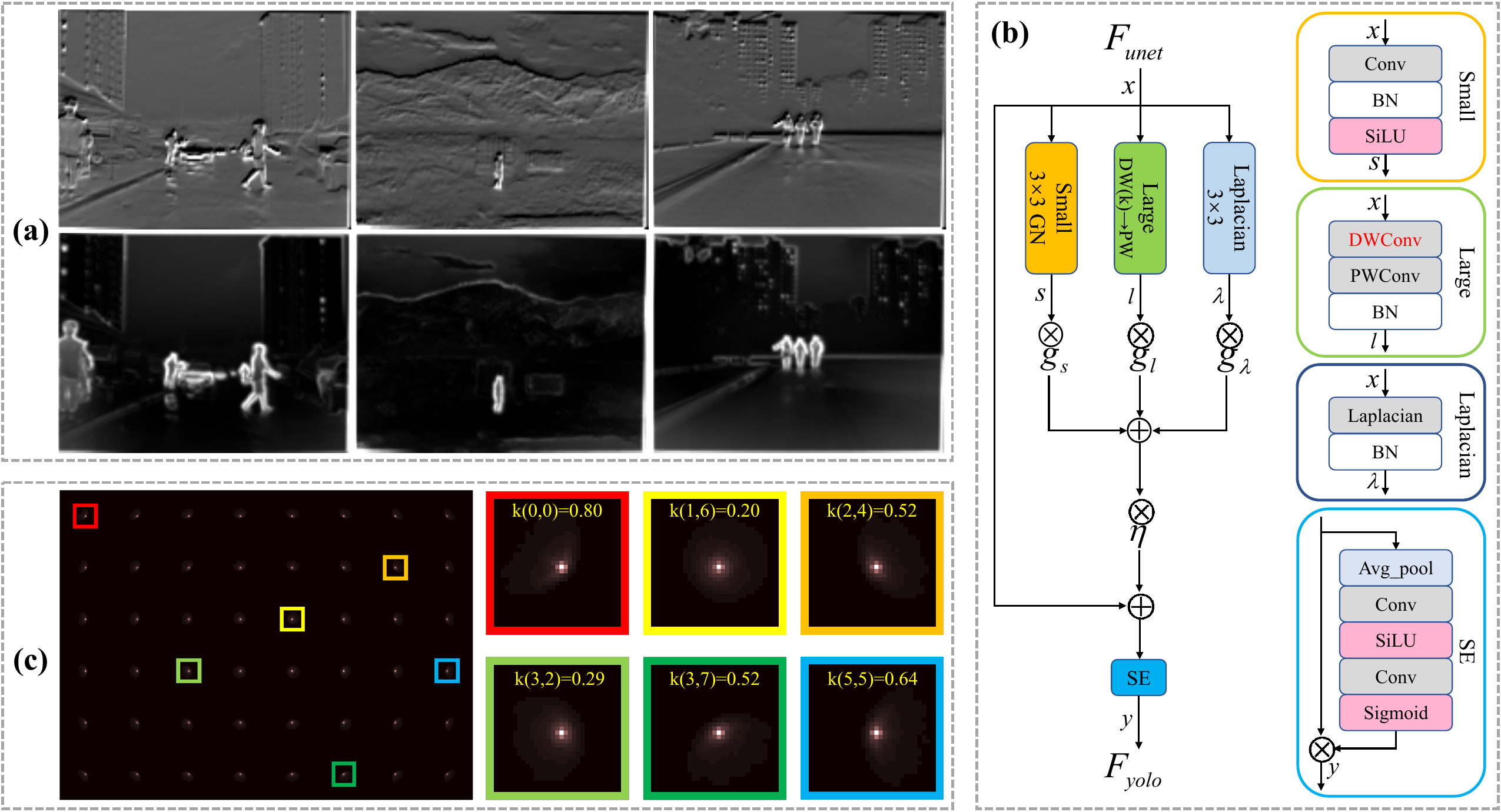}
		\caption{(a) Feature discrepancy between reconstruction and detection modules. (b) Detailed architecture of the PALS-Bridge. (c) The partitioned PSF pattern of the single-lens infrared camera, divided into 48 regions according to the FOV.}
		\label{fig:pals-bridge}
	\end{figure*}

	Establishing an efficient connection between the reconstruction and detection modules is critical to the proposed PDI-Net. However, these two tasks exhibit fundamentally different feature extraction characteristics. Specifically, the U-Net-based reconstruction task is fidelity-driven, treating all regions and pixels uniformly to restore fine-grained texture details across the entire image, including redundant background information. In contrast, the YOLO-based detection backbone is semantics-driven, where feature responses are sparse and concentrated on target regions while actively suppressing most background details. As shown in Fig.~\ref{fig:pals-bridge}(a), the reconstruction and detection tasks present distinct feature distribution patterns, highlighting a clear representational gap between them. This discrepancy confirms that direct feature transfer is inefficient and necessitates a dedicated alignment mechanism.
	
	The proposed PALS-Bridge is designed to mitigate the representational gap between dense, low-level image features and high-level semantic features. The core design principle incorporates PSF characteristics from different field-of-view (FOV) positions as physical priors, adaptively adjusting the weights of multiscale feature extraction branches according to the degree of spot dispersion on the focal plane. Intuitively, in regions with pronounced blur, where optical diffusion is more severe, features are extracted using a larger receptive field to preserve global structural information, whereas in regions with mild blur, smaller receptive fields are emphasized to enhance fine-grained textures and edge sharpness. The detailed architecture of the proposed PALS-Bridge is illustrated in Fig.~\ref{fig:pals-bridge}(b), with a squeeze-and-excitation (SE) submodule for lightweight channel reweighting. Given an input feature map $x \in \mathbb{R}^{B\times C\times H\times W}$, where $x$ corresponds to the encoder feature $F_{\textit{unet}}$, PALS-Bridge processes $x$ through three parallel convolutional branches with different kernel sizes, enabling a balanced extraction of fine-grained local details and coarse global contextual features. The three branches, illustrated in Fig.~\ref{fig:pals-bridge}(b) and defined by Eq.~(\ref{eq:pals-paths}), take the feature map $x$ from the U-Net as their shared input and produce the outputs $s$, $l$, and $\lambda$, respectively. After appropriate scaling, the three outputs are merged with $x$ and forwarded to the SE submodule to generate the final output $y$. The weighting coefficients of the three branches, denoted as $g_s$, $g_l$, and $g_{\lambda}$, are adaptively adjusted according to the PSF characteristics across the FOV.
	
	\begin{equation}
		\begin{aligned}
			s \;&=\; f_{\text{small}}(x)\quad\text{(grouped $3{\times}3$ conv + norm + SiLU)} \\
			l \;&=\; \mathrm{BN}_{l}\!\big(\mathrm{PW}(\mathrm{DW}_{k}(x))\big)\quad\text{(depthwise $15{\times}15$ $\rightarrow$ pointwise)} \\
			\lambda \;&=\; \mathrm{BN}_{\lambda}\!\big(x \ast K_{\lambda}\big)\quad\text{(fixed $3{\times}3$ Laplacian)}
		\end{aligned}
		\label{eq:pals-paths}
	\end{equation}
	
	As shown in Fig.~\ref{fig:pals-bridge}(c), the $640 \times 480$ pixel array, obtained from the single-lens infrared system, is divided into 48 regions of $6 \times 8$ according to the FOV. For each region, the energy centroid and the root-mean-square (RMS) radius of energy around the centroid are calculated based on the experimentally calibrated PSFs, yielding a scalar blur factor. These blur factors are normalized to the range $[0,1]$, arranged on a $6 \times 8$ grid, and bilinearly upsampled to $H \times W$ to form the PSF-aware blur index map $k(h,w)$. The base gating coefficients $\bar g_s$, $\bar g_l$, and $\bar g_{\lambda}$ are then spatially modulated according to the PSF-aware blur index map, yielding the spatially varying gates defined as follows:
	\begin{equation}
		\begin{aligned}
			g_s(h,w) &= \operatorname{clip}_{[0,1]}\!\big(\bar g_s + \alpha_s(1-k(h,w))\big), \\
			g_l(h,w) &= \operatorname{clip}_{[0,1]}\!\big(\bar g_l + \alpha_l\,k(h,w)\big), \\
			g_{\lambda}(h,w) &= \operatorname{clip}_{[0,1]}\!\big(\bar g_{\lambda} + \alpha_{\lambda}(1-k(h,w))\big),
		\end{aligned}
		\label{eq:pals-psf-gates}
	\end{equation}
	where $\alpha_s$, $\alpha_l$, and $\alpha_{\lambda}$ are learnable modulation strengths. Intuitively, the module relies more on the large-kernel path in regions with pronounced blur ($k \uparrow$), and favors the small-kernel and Laplacian-enhanced paths in sharper regions ($k \downarrow$). For stable training, the gates are parameterized as $\bar g_s=\mathrm{sigmoid}(\theta_s)$, $\bar g_l=\mathrm{sigmoid}(\theta_l)$, and $\bar g_{\lambda}=\mathrm{sigmoid}(\theta_{\lambda})$, with the biases empirically initialized to encode a conservative prior over branch importance, while remaining fully learnable.
	
	The gated outputs are fused with a scaled residual connection and lightweight channel reweighting. The overall output of PALS-Bridge is formulated as
	\begin{equation}
		y \;=\; \mathrm{BN}_{\text{out}}\!\Big(\mathrm{SE}\big(x + \eta[ g_s \odot s + g_l \odot l + g_{\lambda} \odot \lambda ]\big)\Big),
		\label{eq:pals-fusion}
	\end{equation}
	where $\eta \in (0,1)$ is a learnable scaling factor that stabilizes early training, $\odot$ is the broadcasted element-wise product, and $\mathrm{BN}_{\text{out}}$ is the final normalization. The PSF prior thus embeds field-dependent physical knowledge with negligible additional inference overhead. PALS-Bridge adaptively modulates spatial frequency responses across the FOV, preserving localization cues in well-focused regions while aggregating robust contextual information in optically blurred areas.
	
	\subsection{Integrated optimization}
	To maximize the optimization efficiency of the integrated framework for image reconstruction and object detection, a joint optimization strategy across multiple modules is employed. This collaborative training enables efficient feature interaction between the two tasks, allowing the reconstruction module to generate feature representations that directly benefit the detector and achieve an optimal balance between convergence speed and detection performance. To realize this joint optimization, we define a total loss function ${\cal L}_{total}$, as shown in Eq.~(\ref{eq:total-loss}), which harmonizes the objectives of both tasks.
	\begin{equation}
		\mathcal{L}_{total} = hyp_{rec} \mathcal{L}_{rec} + hyp_{det} \mathcal{L}_{det},
		\label{eq:total-loss}
	\end{equation}
	where $hyp_{rec}$ and $hyp_{det}$ are weighting factors that regulate the relative contributions of the reconstruction loss ${\cal L}_{rec}$ and the detection loss ${\cal L}_{det}$. It is worth noting that although the network performs full image reconstruction during training, only the feature-sharing layers of the reconstruction module are utilized during inference. Since the reconstruction process is computationally intensive, this design substantially reduces inference time and significantly enhances the feasibility of real-time deployment on edge devices.
	
\begin{figure*}[!t]
	\centering
	\includegraphics[width=\textwidth]{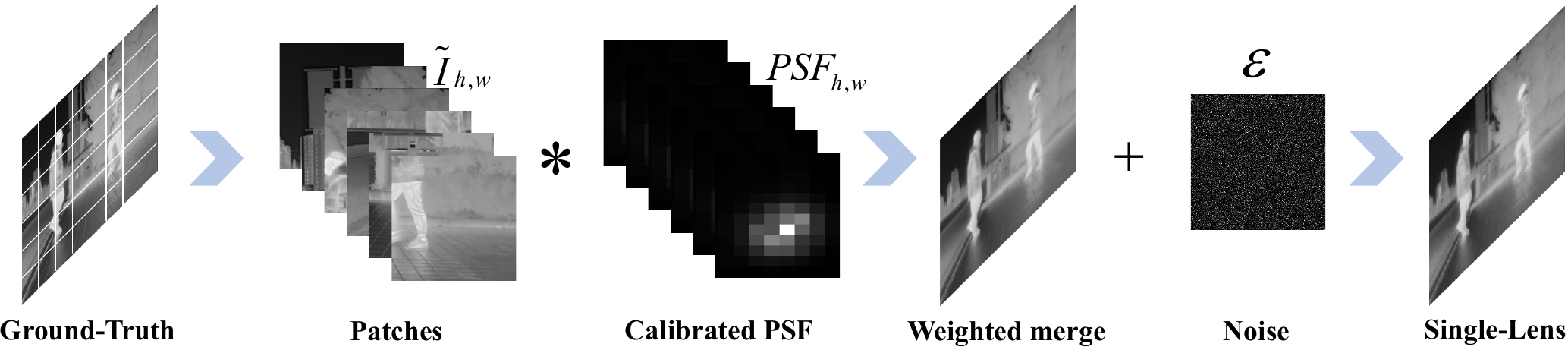}
	\caption{Simulation-based dataset generation process for single-lens infrared computational imaging cameras.}
	\label{fig:data-pipeline}
\end{figure*}

\subsection{Dataset generation}
\label{sec:dataset-generation}
The generation of paired blurred and clear infrared images that incorporate both target annotations and single-lens characteristics is essential for training. However, existing infrared datasets lack the specific optical characteristics of single-lens computational imaging systems and therefore cannot be directly utilized for this purpose. To embed physics-informed priors into the training data, we simulate single-lens degradations using calibrated PSFs and detector noise models. The degradation process accounts for residual aberrations, optical blur, and spatial distortions that are intrinsic to single-lens designs.
	
To obtain physically realistic degradation kernels, the PSFs used in this work are derived from an optical simulation model of the single-lens infrared imaging system. Specifically, the optical system is constructed in Zemax OpticStudio and the residual wavefront aberrations are exported through the ZOSAPI interface. For a given field position and wavelength, the complex pupil function can be expressed as
\begin{equation}
	U(x,y) = A(x,y)e^{i2\pi W(x,y)}.
\end{equation}
where $A(x,y)$ represents the pupil aperture function, $W(x,y)$ denotes the wavefront aberration obtained from optical simulation, and $i=\sqrt{-1}$ is the imaginary unit, with the exponential term describing the phase modulation of the complex optical field. Based on scalar diffraction theory, the corresponding PSF is computed as the squared magnitude of the Fourier transform of the complex pupil function:
\begin{equation}
	\mathrm{PSF}(u,v) = \left| \mathcal{F}\{U(x,y)\} \right|^2.
\end{equation}
	
To account for the broadband infrared imaging characteristics of the system, PSFs are generated at multiple wavelengths within the spectral range of $8$--$12~\mu$m and then spectrally averaged to obtain a broadband kernel. In addition, since the imaging performance varies with the field position due to off-axis aberrations, PSFs are sampled on a grid of field points across the sensor plane. Finally, the simulated PSFs are resampled according to the detector pixel size to ensure consistency with the physical imaging process. This procedure produces a spatially varying PSF set $\{K_{m,n}\}$ that accurately reflects the optical response of the single-lens imaging system.
	
	As illustrated in Fig.~\ref{fig:data-pipeline}, the degradation simulation consists of three primary operations: cropping, convolution, and noise addition. First, the ground truth image $I_{gt} \in \mathbb{R}^{M \times N}$ is divided into patches $\widetilde I \in \mathbb{R}^{p \times p}$, where $M$ and $N$ are integer multiples of $p$. Each patch $\widetilde I$ is then convolved with the calibrated PSF $K \in \mathbb{R}^{d_{\mathrm{psf}} \times d_{\mathrm{psf}}}$, and the resulting outputs are recombined using a weighting function. Finally, Gaussian noise $\varepsilon$ is added, with its intensity controlled by a scaling parameter $q$. To prevent numerical instability, the resulting data are clamped to the range $(10^{-20},\,1.0)$.
	
	Mathematically, the calibrated PSF degradation process can be expressed as follows. Let $M_p=M/p$ and $N_p=N/p$ denote the numbers of patches along the two spatial dimensions:
	\begin{equation}
		\widehat{I} = \sum\limits_{m=0}^{M_p-1} \sum\limits_{n=0}^{N_p-1} W_{m,n} \odot \left( \widetilde{I}_{m,n} * K_{m,n} \right),
		\label{eq:psf-degradation}
	\end{equation}
	where $\widetilde I_{m,n}$ represents the $m$-th and $n$-th patch, $K_{m,n}$ means the corresponding PSF, $*$ refers to the 2D convolution operation, which models the spatially varying degradation across the image by applying field-dependent PSFs. $W_{m,n}$ denotes the weighting matrix of the corresponding patch, and $\odot$ denotes element-wise multiplication. At this stage, the PSF-degraded image $\widehat{I}$ is obtained. In fact, the degradation of single-lens is not limited to calibrated PSF, but also includes sensor noise. Therefore, we use Gaussian noise to simulate this process:
	\begin{equation}
		I_{raw} = \min \left( \max \left( \left( \left\lfloor \frac{\widehat{I}}{q} \right\rfloor + \varepsilon \right) \times q, 10^{-20} \right), 1.0 \right),
		\label{eq:noise-model}
	\end{equation}
	where $\widehat{I}$ is the PSF-degraded and recombined image. To control the noise level, a scaling factor $q$ is introduced, and $\varepsilon \sim \mathcal{N}(0, \sigma^2)$ models additive Gaussian noise. In Eq.~(\ref{eq:noise-model}), the clamp operation is represented by the combination of min and max functions.
	
	\section{Experiments}
	\subsection{Experimental settings}
	1) Camera: The compact single-lens infrared camera used in this study is a self-developed system featuring a hybrid refractive--diffractive optical structure with an end-to-end co-design of the optical system and neural network~\cite{xing2025real}. The camera operates within the 8--12~$\mu$m spectral range, offering a focal length of 70~mm and an F-number of 1.0. It employs an uncooled infrared detector with a spatial resolution of $640\times480$ pixels. The system also integrates an RK3588 NPU chip, which in our previous design and preliminary studies~\cite{Wang2025} was mainly dedicated to running optimized image reconstruction algorithms. In this work, the same camera and its calibrated PSF are employed to experimentally validate the proposed integrated reconstruction--detection framework.
	
	2) Dataset: The dataset used in this study is derived from the \textnormal{M}\textsuperscript{3}\textnormal{FD} benchmark~\cite{liu2022target} through the degradation process described in Section~\ref{sec:dataset-generation}. The original dataset contains 4,200 pairs of infrared and visible images. Since the single-lens infrared camera used in this work operates at a resolution of $640\times480$ pixels, whereas most images in the \textnormal{M}\textsuperscript{3}\textnormal{FD} dataset have a resolution of $1024\times768$, we select the annotated infrared frames at $1024\times768$ and uniformly rescale both the images and bounding boxes to $640\times480$. This procedure preserves annotation fidelity and ensures consistency with the target deployment resolution. To further enhance the robustness of experimental evaluation, the FLIR\_ADAS\_v2 dataset is also incorporated. A detailed dataset comparison is summarized in Table~\ref{tab:dataset-overview}.
	
	\begin{table*}[!t]
		\centering
		\footnotesize
		\caption{Dataset Overview of the M\textsuperscript{3}FD and FLIR\_ADAS\_v2}
		\label{tab:dataset-overview}
		\begin{tabular*}{0.9\textwidth}{@{\extracolsep{\fill}}lrrcc}
			\toprule
			Name & Train & Test & Size & Class \\
			\midrule
			M\textsuperscript{3}FD & 3144 & 782 & 640$\times$480 & 6 \\
			FLIR\_ADAS\_v2 & 10742 & 1144 & 640$\times$512 & 16 \\
			\bottomrule
		\end{tabular*}
	\end{table*}
	
	3) Evaluation Metrics: To comprehensively evaluate the inference time and object detection performance, \textbf{Speed} and several reference-based metrics are adopted, including \textbf{Precision}, \textbf{Recall}, mean Average Precision (mAP) at Intersection over Union (IoU) = 0.5 (\textbf{mAP@0.5}), and mAP averaged over IoU thresholds from 0.5 to 0.95 (\textbf{mAP@0.5:0.95}). \textbf{Speed} denotes inference time per image in milliseconds, where smaller values indicate faster inference. Higher values of these accuracy metrics approaching 1 indicate superior detection quality and robustness.
	
	4) Implementation Details: The proposed integrated network for reconstruction and detection is implemented using the PyTorch framework and trained on an NVIDIA GeForce RTX 4060 GPU. The scale variable $q$ and standard deviation $\sigma$ are set to 90 and 0.0003, respectively. The training process employs the SGD optimizer with a learning rate of 0.01, a batch size of 16, and a total of 300 training epochs. The hyperparameters are configured as follows: the reconstruction loss weight $hyp_{rec}$ is set to 0.01, the detection loss weight $hyp_{det}$ to 1.0, the bounding box localization loss weight $hyp_{box}$ to 0.05, the objectness loss weight $hyp_{obj}$ to 1.0, and the classification loss weight $hyp_{cls}$ to 0.5. The residual-scaling factor is set to $\eta=0.2$. The global gate biases are empirically initialized as $\theta_s=-3.0$, $\theta_l=-2.0$, and $\theta_{\lambda}=-4.0$.
	
	\subsection{Quantitative comparison}
	\begin{table*}[!t]
		\centering
		\footnotesize
		\caption{Quantitative Comparison of Different Infrared Imaging Strategies}
		\label{tab:strategy-comparison}
		\begin{tabular*}{\textwidth}{@{\extracolsep{\fill}}lcccccc}
			\toprule
			Strategy & SNR & Speed (ms) & Precision & Recall & mAP@0.5 & mAP@0.5:0.95 \\
			\midrule
			\multirow{2}{*}{Traditional imaging} & High & \multirow{2}{*}{4.53} & 0.861 & 0.734 & 0.808 & 0.513 \\
			& Low &  & 0.826 & 0.742 & 0.802 & 0.506 \\
			\midrule
			\multirow{2}{*}{Rec+Det} & High & \multirow{2}{*}{58.41} & 0.881 & 0.730 & 0.814 & 0.517 \\
			& Low &  & 0.852 & 0.716 & 0.794 & 0.495 \\
			\midrule
			\multirow{2}{*}{Rec+Det with pruning} & High & \multirow{2}{*}{34.64} & 0.843 & 0.728 & 0.795 & 0.494 \\
			& Low &  & 0.855 & 0.698 & 0.784 & 0.493 \\
			\midrule
			\multirow{2}{*}{Raw imaging} & High & \multirow{2}{*}{4.60} & 0.725 & 0.505 & 0.575 & 0.340 \\
			& Low &  & 0.786 & 0.400 & 0.499 & 0.286 \\
			\midrule
			\multirow{2}{*}{Our work} & High & \multirow{2}{*}{5.52} & 0.882 & 0.757 & 0.825 & 0.521 \\
			& Low &  & 0.871 & 0.752 & 0.813 & 0.518 \\
			\bottomrule
		\end{tabular*}
	\end{table*}
	
	To directly evaluate the performance of the proposed method, we compared the five infrared imaging strategies illustrated in Fig.~\ref{fig:strategy}. Table~\ref{tab:strategy-comparison} summarizes the quantitative results on the \textnormal{M}\textsuperscript{3}\textnormal{FD} dataset, including inference speed on GPU and detection metrics such as Precision, Recall, mAP@0.5, and mAP@0.5:0.95. For fairness, both the traditional multi-lens and single-lens systems were modeled using their respective PSF characteristics, and degraded datasets were generated using identical procedures.
	
	Among the evaluated imaging strategies, traditional multi-lens imaging achieves the shortest inference time owing to the absence of a reconstruction stage. In contrast, the Rec+Det pipeline introduces substantial computational overhead; even after applying 50\% channel pruning, its inference time remains 34.64~ms, significantly slower than traditional imaging. Similarly, direct detection on raw degraded inputs achieves an inference latency comparable to that of traditional multi-lens imaging. In comparison, the proposed PDI-Net achieves an inference speed of 5.52~ms, approximately matching traditional imaging, and reduces inference time by 84.06\% compared with the \textit{Rec+Det with pruning} strategy, because it leverages shared encoder features rather than fully reconstructed images.

	To assess robustness under varying noise levels, experiments were conducted at two SNR conditions: high SNR (noise scale $q=10$) and low SNR ($q=90$). Under high SNR, traditional imaging surpasses only the pruned Rec+Det pipeline, which is unsurprising given its lack of a learnable recovery stage, while the proposed PDI-Net achieves the highest detection accuracy among all computational imaging strategies. Unsurprisingly, raw imaging performs the worst and degrades markedly, as it lacks both ground truth supervision and physical aberration correction. For the detection task, our strategy follows the same configuration as raw imaging, differing only in the detection model; nevertheless, PDI-Net improves mAP@0.5 by 43.48\% at the cost of only about 1~ms additional inference time. As the SNR decreases, all methods experience performance degradation; however, the Rec+Det pipeline suffers the most pronounced drop. Traditional multi-lens imaging demonstrates the highest stability, followed closely by the proposed PDI-Net. Under the low-SNR setting, PDI-Net further improves mAP@0.5:0.95 by 5.07\% compared with the \textit{Rec+Det with pruning} strategy. This can be attributed to the differing optimization objectives: reconstruction-first pipelines are typically trained for human-perceptual fidelity and tend to oversmooth images under low SNR, thereby suppressing edges and textures critical for detection. In contrast, our integrated approach jointly aligns shared features with the detection objective, enhancing task-relevant spatial frequencies while de-emphasizing visually redundant information.
		
	\begin{figure*}[!b]
		\centering
		\includegraphics[width=\textwidth]{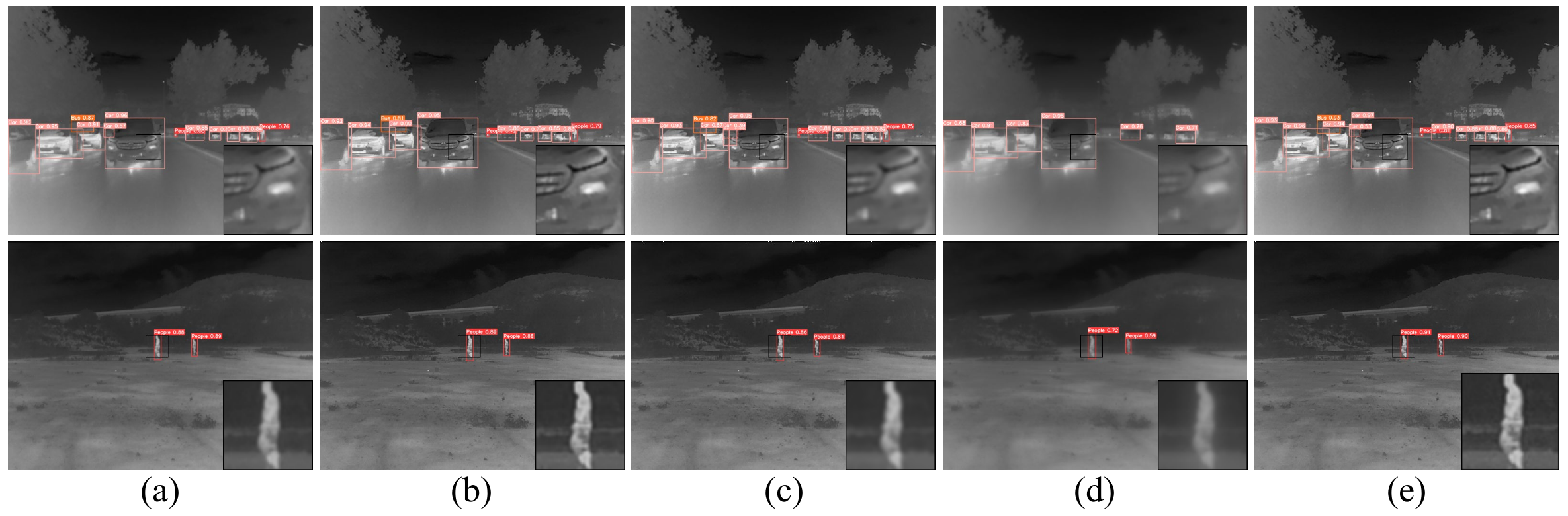}
		\caption{Qualitative comparison of infrared object detection with different imaging strategies. (a) Traditional imaging. (b) Rec+Det. (c) Rec+Det with pruning. (d) Raw imaging. (e) The proposed PDI-Net.}
		\label{fig:qualitative-comparison}
	\end{figure*}
	
	Overall, these results confirm that the proposed PDI-Net achieves an effective balance among real-time efficiency, detection accuracy and lightweight system design, maintaining strong robustness under noise variations. Given that most practical infrared imaging scenarios operate in low-SNR conditions, subsequent experiments are conducted under the low-SNR setting to better reflect real-world application environments.
	
	\subsection{Qualitative comparison}
	\label{sec:qualitative-comparison}

	From a visual perspective, Fig.~\ref{fig:qualitative-comparison} presents the detection outcomes of different imaging strategies.
	
	In terms of perceptual image quality, computational imaging pipelines exhibit performance comparable to traditional multi-lens imaging. Without any learnable recovery module, the traditional approach relies solely on optical correction, which limits noise suppression and mid-to-high frequency restoration. Among the computational imaging strategies in Fig.~\ref{fig:qualitative-comparison}(b), (c) and (e), the full reconstruction network in Fig.~\ref{fig:qualitative-comparison}(b) provides the finest structural detail. Applying 50\% channel pruning introduces minor texture loss, as shown in Fig.~\ref{fig:qualitative-comparison}(c), consistent with reduced model capacity. The integrated framework in Fig.~\ref{fig:qualitative-comparison}(e) achieves comparable structural fidelity to Fig.~\ref{fig:qualitative-comparison}(b) while significantly lowering latency, as it directly consumes shared encoder features rather than performing full image reconstruction. Expectedly, raw imaging exhibits the poorest visual quality due to the absence of both reconstruction algorithms and optical aberration correction.
	
	From the detection perspective, the proposed PDI-Net yields consistently higher confidence scores across all targets. For instance, the orange \textit{Bus} label reaches a confidence of 0.93, compared with 0.87, 0.81, and 0.82 in Fig.~\ref{fig:qualitative-comparison}(a)--(c), respectively. This improvement arises from the joint optimization of reconstruction and detection, which aligns feature extraction with detection objectives rather than human-perceptual quality. Consequently, features enhanced by the reconstruction module are more discriminative and task-oriented, leading to superior detection accuracy even under degraded imaging conditions. In contrast, the Rec+Det pipelines in Fig.~\ref{fig:qualitative-comparison}(b)--(c) prioritize perceptual quality during reconstruction, which may conflict with machine perception and ultimately limit detection robustness compared with the integrated network. Unsurprisingly, the raw imaging strategy yields the worst detection performance, with some targets missed and generally lower confidence scores, due to the absence of both reconstruction algorithms and optical aberration correction.
	
\begin{figure*}[htbp]
		\centering
		\includegraphics[width=\textwidth]{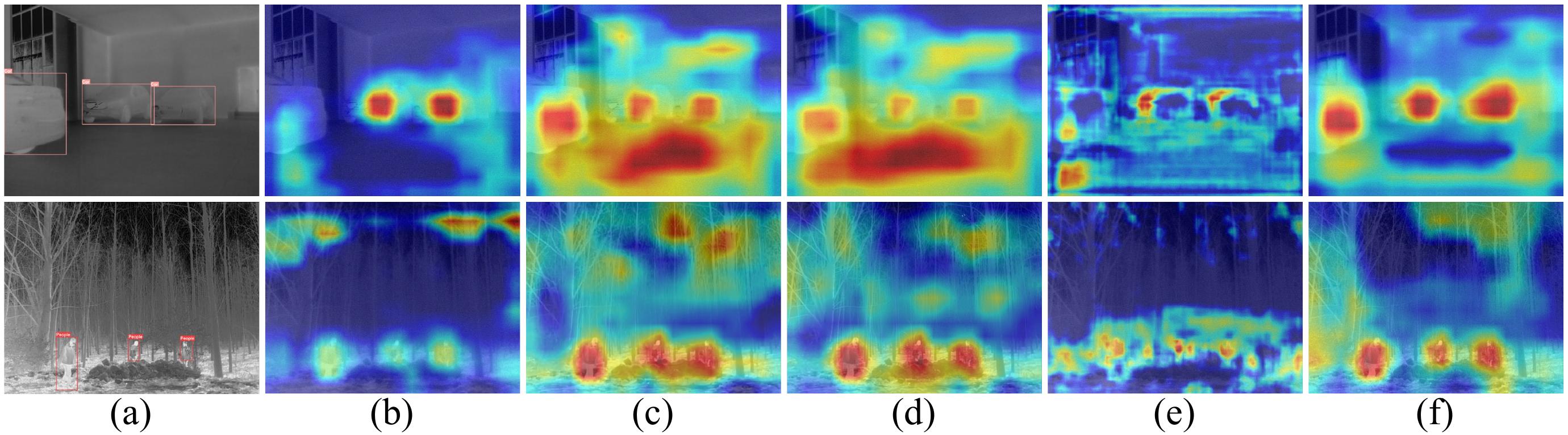}
		\caption{Heatmap comparison of different infrared imaging strategies. (a) Ground truth. (b) Traditional imaging. (c) Rec+Det. (d) Rec+Det with pruning. (e) Raw imaging. (f) The proposed PDI-Net.}
		\label{fig:heatmap-comparison}
	\end{figure*}

	We further provide Grad-CAM~\cite{selvaraju2017grad} visualizations to highlight where each detector concentrates during inference. Fig.~\ref{fig:heatmap-comparison} shows heatmaps extracted from the activation of the third convolutional block in the detection head, with warmer colors indicating stronger attention.
	
	\begin{table*}[!b]
		\centering
		\footnotesize
		\caption{Ablation Experiments on Different Connections Between Reconstruction and Detection}
		\label{tab:ablation-connections}
		\begin{tabular*}{\textwidth}{@{\extracolsep{\fill}}lccccc}
			\toprule
			Strategy & Speed (ms) & Precision & Recall & mAP@0.5 & mAP@0.5:0.95 \\
			\midrule
			Strategy.1 red & 5.52 & 0.867 & 0.694 & 0.787 & 0.484 \\
			Strategy.1 yellow & 5.68 & 0.852 & 0.725 & 0.794 & 0.487 \\
			Strategy.1 black & 5.54 & 0.859 & 0.741 & 0.807 & 0.515 \\
			Strategy.2 green & 5.66 & 0.833 & 0.722 & 0.786 & 0.483 \\
			Strategy.2 red & 5.84 & \textbf{0.88} & \underline{0.742} & \underline{0.811} & 0.516 \\
			Strategy.2 yellow & 5.44 & 0.859 & 0.734 & 0.807 & \textbf{0.52} \\
			Strategy.2 black & \textbf{5.2} & 0.845 & 0.705 & 0.78 & 0.479 \\
			Strategy.3 yellow & 5.52 & \underline{0.871} & \textbf{0.752} & \textbf{0.813} & \underline{0.518} \\
			Strategy.3 red & 5.42 & 0.867 & 0.706 & 0.78 & 0.477 \\
			Strategy.3 black & \textbf{5.2} & 0.871 & 0.691 & 0.774 & 0.473 \\
			\bottomrule
		\end{tabular*}
	\end{table*}
	
	In Fig.~\ref{fig:heatmap-comparison}(a), the ground truth layout indicates the ideal regions of interest that the model should attend to. In Fig.~\ref{fig:heatmap-comparison}(b), attention largely overlaps the targets but remains relatively diffuse, which is consistent with residual noise and limited learnable restoration. In Fig.~\ref{fig:heatmap-comparison}(c), attention over the \textit{People} target is concentrated, suggesting that reconstruction improves signal quality, yet part of the \textit{Car} attention drifts to a background area below the object, revealing a mismatch between features emphasized during reconstruction and those required for detection. A similar mismatch appears in Fig.~\ref{fig:heatmap-comparison}(d), which depicts the pruned variant and is likely exacerbated by reduced model capacity. Fig.~\ref{fig:heatmap-comparison}(e) reveals that raw imaging struggles to concentrate attention on the targets, stemming from the absence of ground truth supervision and task-relevant guidance. By contrast, Fig.~\ref{fig:heatmap-comparison}(f) exhibits the most compact and target-aligned attention, avoiding off-target responses seen in (c)--(d). This behavior aligns with the design of the dual-integrated framework, where the detector directly consumes shared encoder features that are optimized jointly with detection, favoring task-relevant structures over purely perceptual fidelity.

	\subsection{Ablation study}
	\label{sec:ablation-study}

\begin{figure*}[!t]
		\centering
		\includegraphics[width=\textwidth]{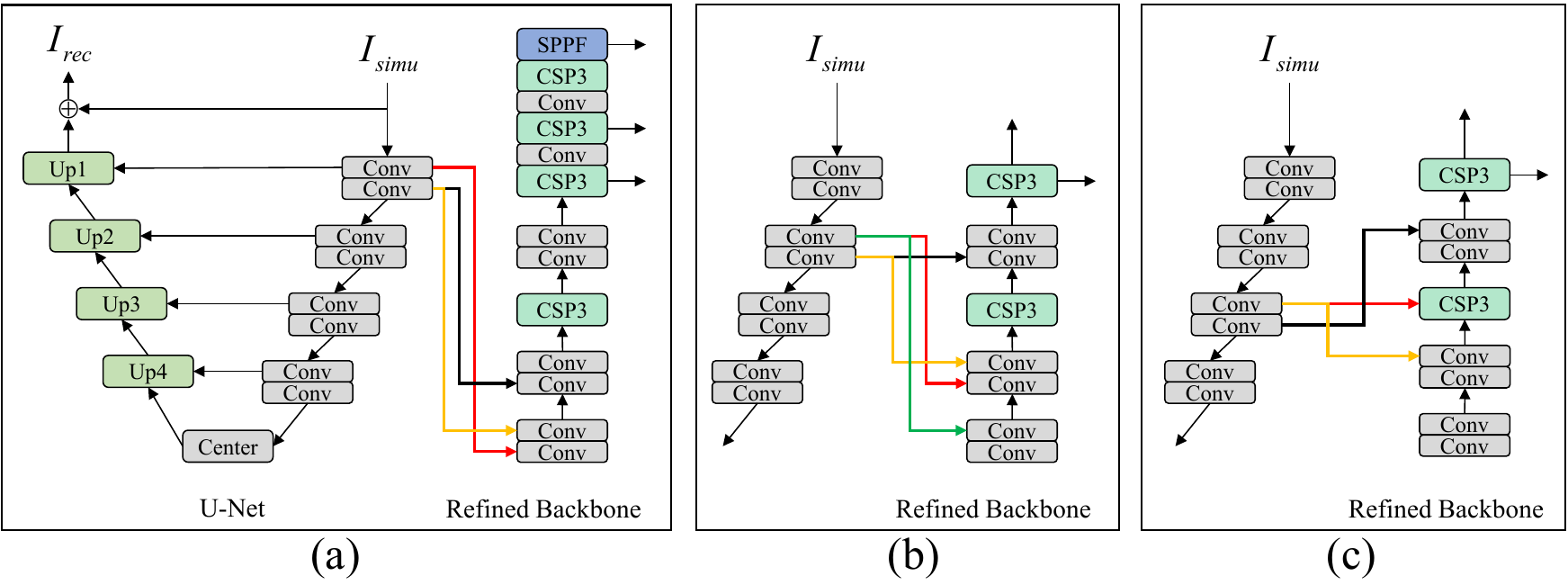}
		\caption{Ablation study and sub-connection methods of the feature-sharing layer. (a) Strategy.1: the first ConvBlock is used as the feature-sharing layer. (b) Strategy.2: the first two ConvBlocks are used as feature-sharing layers. (c) Strategy.3: the first three ConvBlocks are used as feature-sharing layers.}
		\label{fig:ablation-sharing}
	\end{figure*}

	In this section, we conduct ablation studies on the feature-sharing configuration and the design of the PALS-Bridge module to assess their impact on detection performance in the proposed dual-integrated framework. Experiments are performed on the M\textsuperscript{3}FD dataset. Fig.~\ref{fig:ablation-sharing} illustrates the three sharing strategies and their sub-connections, while Tables~\ref{tab:ablation-connections} and~\ref{tab:ablation-bridge} summarize the quantitative results.
	
	\textit{1) Ablation of U-Net Feature-Sharing:} Fig.~\ref{fig:ablation-sharing}(a) presents the improved U-Net architecture, consistent with the detailed U-Net architecture shown in Supplementary Fig. S1. To clarify the connection strategies between the reconstruction and detection modules, the U-Net is shown as a horizontal mirror in Fig.~\ref{fig:ablation-sharing}(a); (b) and (c) provide simplified schematics of the U-Net and the refined backbone, thereby highlighting the detailed structure of the sub-connections.

	\begin{table*}[!b]
		\centering
		\footnotesize
		\caption{Ablation Study of PALS-Bridge and Variants}
		\label{tab:ablation-bridge}
		\begin{tabular*}{0.85\textwidth}{@{\extracolsep{\fill}}lcccc}
			\toprule
			Method & Precision & Recall & mAP@0.5 & mAP@0.5:0.95 \\
			\midrule
			Baseline (w.o. bridge) & 0.871 & \textbf{0.752} & 0.813 & 0.518 \\
			L-Bridge & 0.844 & 0.733 & 0.818 & 0.559 \\
			S-Bridge & 0.855 & 0.724 & 0.816 & 0.552 \\
			LS-Bridge & 0.853 & \underline{0.749} & 0.827 & \underline{0.568} \\
			PALS-Bridge (w.o. SE) & \underline{0.876} & 0.738 & \underline{0.831} & \textbf{0.570} \\
			PALS-Bridge (w.o. Rec) & 0.838 & 0.695 & 0.765 & 0.469 \\
			PALS-Bridge & \textbf{0.879} & 0.746 & \textbf{0.833} & \textbf{0.570} \\
			\bottomrule
		\end{tabular*}
	\end{table*}
	
	To ensure a fair comparison across connection strategies, the original YOLO backbone is lightly modified by expanding the initial three convolution layers to six, which provides more flexible attachment points for external features while keeping resolution and channel alignment consistent with the encoder. As illustrated in Fig.~\ref{fig:ablation-sharing}, Strategy~1, Strategy~2 and Strategy~3 tap features after ConvBlock~1, ConvBlock~2 and ConvBlock~3. Each strategy uses four color-coded sub-connections, namely red, yellow, green, and black, to probe depth and path sensitivity. The results in Table~\ref{tab:ablation-connections} show that sub-connections that bypass CSP3 modules, for example the black sub-connection in Strategy~2 and the black sub-connection in Strategy~3, lead to noticeable accuracy drops, which indicates that preserving the internal transformation within the backbone is critical. Among all configurations, the yellow sub-connection in Strategy~3 delivers the best trade-off by producing the highest mAP@0.5 and Recall while maintaining competitive mAP@0.5:0.95 and Precision at similar latency, therefore adopted in the final model. Deeper sharing beyond ConvBlock~3, referred to as ConvBlock~4, is not considered since matching those deeper features requires bypassing CSP3 modules, which led to pronounced performance degradation.

	\textit{2) Ablation of PALS-Bridge Structure:} Table~\ref{tab:ablation-bridge} evaluates the proposed bridging block. Adding either the small \textit{S} or the large \textit{L} branch individually improves the baseline because the small branch enhances fine-grained detail while the large branch captures long-range context. The Laplacian branch uses the same $3\times3$ kernel as the small branch and is therefore merged into the small branch. Combining small and large branches \textit{LS} further increases accuracy, confirming the benefit of multiscale receptive fields. Introducing the physics-aware PSF-gated path \textit{PA} adaptively reweights the spatial frequency content according to field-dependent blur, emphasizing the large-kernel path in blur-dominant regions while favoring the small-kernel and Laplacian paths in sharper areas; this variant corresponds to PALS-Bridge (no SE) and delivers the strongest improvement in localization. Adding SE reweighting achieves the best overall balance; the full variant is reported as PALS-Bridge and adopted as the default configuration. Removing the reconstruction loss (w.o. Rec) significantly degrades mAP, underscoring the critical role of our joint optimization strategy in guiding feature extraction.

	\subsection{Comparison with traditional imaging frameworks}
	\label{sec:comparison-traditional-frameworks}
	To demonstrate the necessity of a dedicated architecture for single-lens computational imaging, we compare the proposed PDI-Net with several representative joint reconstruction--detection frameworks originally developed for traditional imaging systems. These networks include: Image-Adaptive YOLO~\cite{IA-YOLO}, designed for detection in adverse weather and low-light conditions; DeepDenoising~\cite{DeepDenoising}, which examines the relationship between denoising and high-level vision tasks; and ESOD~\cite{ESOD}, which focuses on small-object detection in high-resolution imagery. We include FLIR\_ADAS\_v2 in addition to the \textnormal{M}\textsuperscript{3}\textnormal{FD} dataset to evaluate generalization, and we report not only detection accuracy, but also parameter counts, FLOPs, and inference speed on the 640$\times$480 version of \textnormal{M}\textsuperscript{3}\textnormal{FD}.
	
	\begin{table*}[!t]
		\centering
		\footnotesize
		\caption{Quantitative Comparison with Traditional Imaging Frameworks on the M\textsuperscript{3}FD Dataset}
		\label{tab:comparison-m3fd}
		\begin{tabular*}{\textwidth}{@{\extracolsep{\fill}}lccccccc}
			\toprule
			Method & Precision & Recall & mAP@0.5 & mAP@0.5:0.95 & Params (M) & FLOPs (G) & Speed (ms) \\
			\midrule
			IA-YOLO & 0.776 & 0.585 & 0.697 & 0.399 & 61.8 & 117.4 & 28.1 \\
			DeepDenoising & 0.828 & 0.578 & 0.723 & 0.478 & \textbf{7.5} & 92.1 & \underline{10.1} \\
			ESOD & \underline{0.833} & \underline{0.630} & \underline{0.756} & \underline{0.485} & 13.6 & \textbf{21.4} & 10.6 \\
			Ours & \textbf{0.879} & \textbf{0.746} & \textbf{0.833} & \textbf{0.570} & \underline{9.2} & \underline{47.8} & \textbf{5.8} \\
			\bottomrule
		\end{tabular*}
	\end{table*}
		
	\begin{figure*}[!t]
		\centering
		\includegraphics[width=\textwidth]{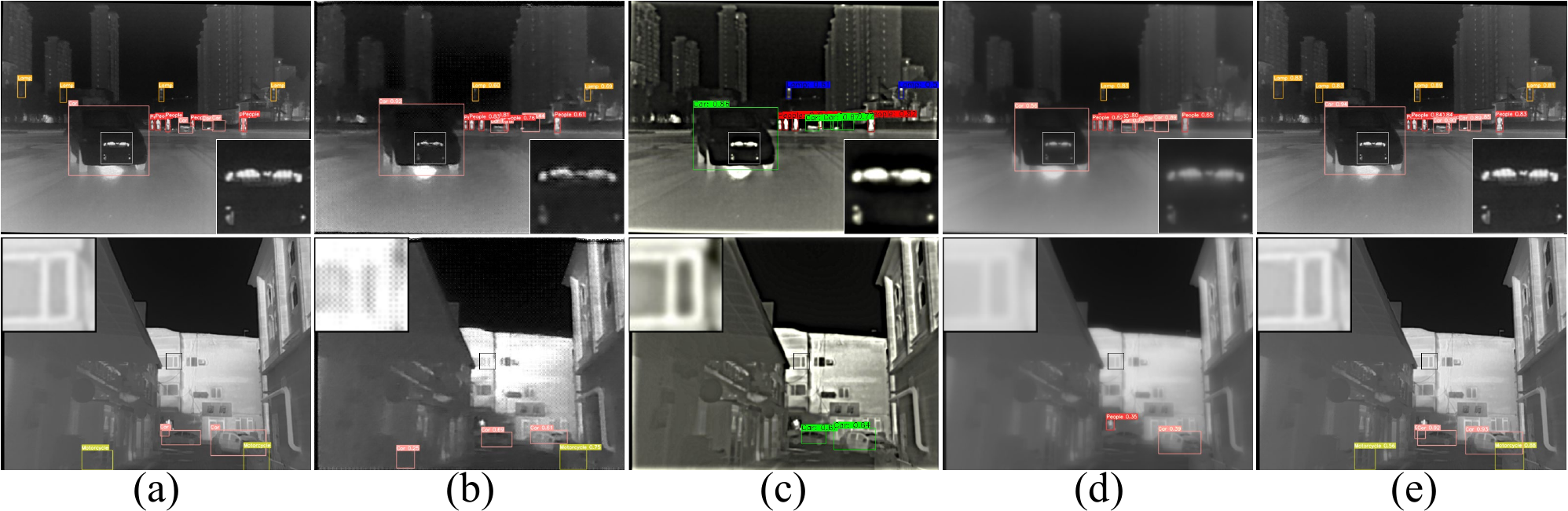}
		\caption{Qualitative comparison with traditional imaging frameworks on the \textnormal{M}\textsuperscript{3}\textnormal{FD} dataset: (a) Ground truth; (b) DeepDenoising; (c) IA-YOLO; (d) ESOD; (e) Ours.}
		\label{fig:comparison-m3fd}
	\end{figure*}
		
	\begin{figure*}[!b]
		\centering
		\includegraphics[width=\textwidth]{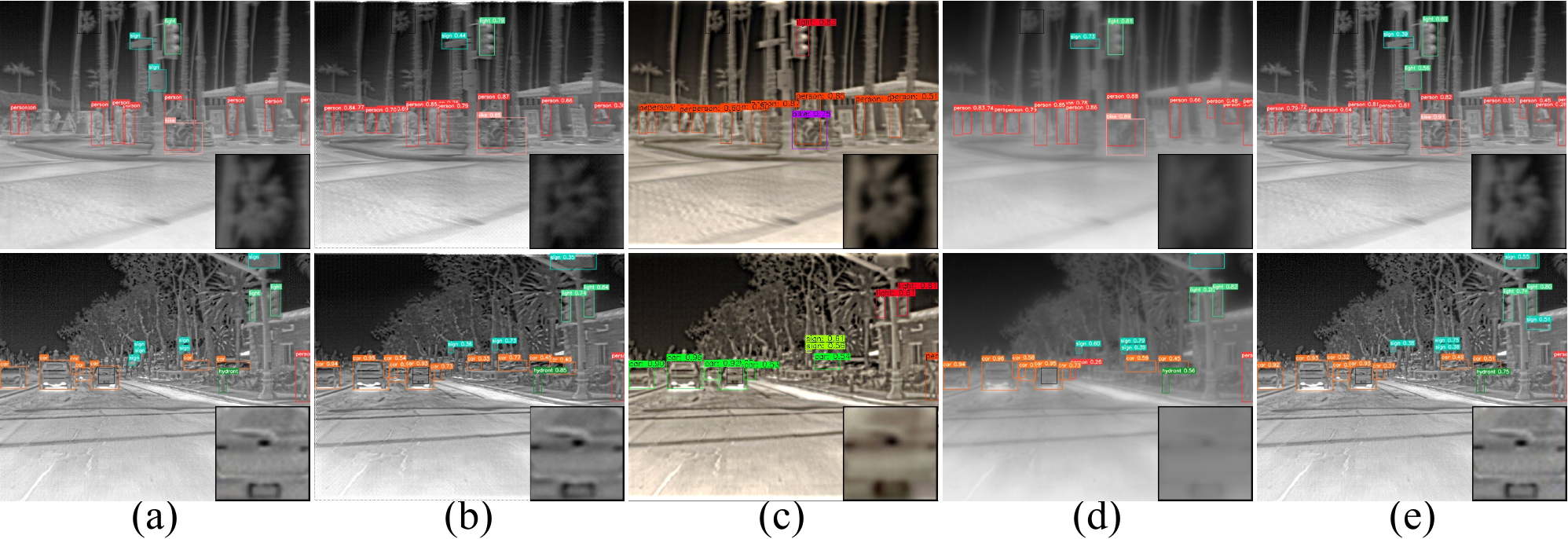}
		\caption{Qualitative comparison with traditional imaging frameworks on the FLIR\_ADAS\_v2 dataset: (a) Ground truth; (b) DeepDenoising; (c) IA-YOLO; (d) ESOD; (e) Ours.}
		\label{fig:comparison-flir}
	\end{figure*}
			
	\begin{table*}[!t]
		\centering
		\footnotesize
		\caption{Quantitative Comparison with Traditional Imaging Frameworks on the FLIR\_ADAS\_v2 Dataset}
		\label{tab:comparison-flir}
		\begin{tabular*}{0.8\textwidth}{@{\extracolsep{\fill}}lcccc}
			\toprule
			Method & Precision & Recall & mAP@0.5 & mAP@0.5:0.95 \\
			\midrule
			IA-YOLO & 0.557 & 0.298 & 0.432 & 0.212 \\
			DeepDenoising & 0.571 & \underline{0.394} & 0.479 & 0.304 \\
			ESOD & \underline{0.588} & \textbf{0.412} & \underline{0.496} & \underline{0.315} \\
			Ours & \textbf{0.599} & \textbf{0.412} & \textbf{0.506} & \textbf{0.322} \\
			\bottomrule
		\end{tabular*}
	\end{table*}
	
	Table~\ref{tab:comparison-m3fd} reports the quantitative results on the \textnormal{M}\textsuperscript{3}\textnormal{FD} dataset. The proposed PDI-Net achieves the best overall performance, surpassing the next-best ESOD by a clear margin, while using substantially fewer parameters than IA-YOLO and delivering the fastest inference among all methods. We attribute this gain to PALS-Bridge, which narrows the gap between pixel-aligned low-level textures and detector-oriented semantics, thereby stabilizing the interface between reconstruction features and the detection head. The comparison also reflects a distribution mismatch: competing detectors are not tailored to single-lens, PSF-degraded infrared imagery and lack physics-aware priors, which likely contributes to their lower scores on our benchmark. Fig.~\ref{fig:comparison-m3fd} further illustrates these trends: in the first row, only the proposed PDI-Net correctly detects the \textit{Lamp}; in the second row, which contains a small, low-contrast \textit{Motorcycle}, PDI-Net is the only method that produces a correct detection and also yields higher confidence on other targets. These observations indicate that the three-branch PALS-Bridge effectively handles small and low-contrast objects. Additionally, the zoomed insets demonstrate that the proposed PDI-Net most closely approximates the ground truth, whereas DeepDenoising and IA-YOLO are inferior, and ESOD lacks a reconstruction process.
			
	\begin{table*}[!t]
		\centering
		\footnotesize
		\caption{Scalability of the Proposed Framework on YOLO Family}
		\label{tab:scalability-yolo}
		\begin{tabular*}{0.8\textwidth}{@{\extracolsep{\fill}}lcccc}
			\toprule
			Method & Precision & Recall & mAP@0.5 & mAP@0.5:0.95 \\
			\midrule
			YOLOv5s-based & \textbf{0.879} & 0.746 & 0.833 & 0.570 \\
			YOLOv8s-based & 0.878 & \textbf{0.754} & \textbf{0.853} & \textbf{0.626} \\
			\bottomrule
		\end{tabular*}
	\end{table*}
										
	\begin{figure*}[!b]
		\centering
		\includegraphics[width=\textwidth]{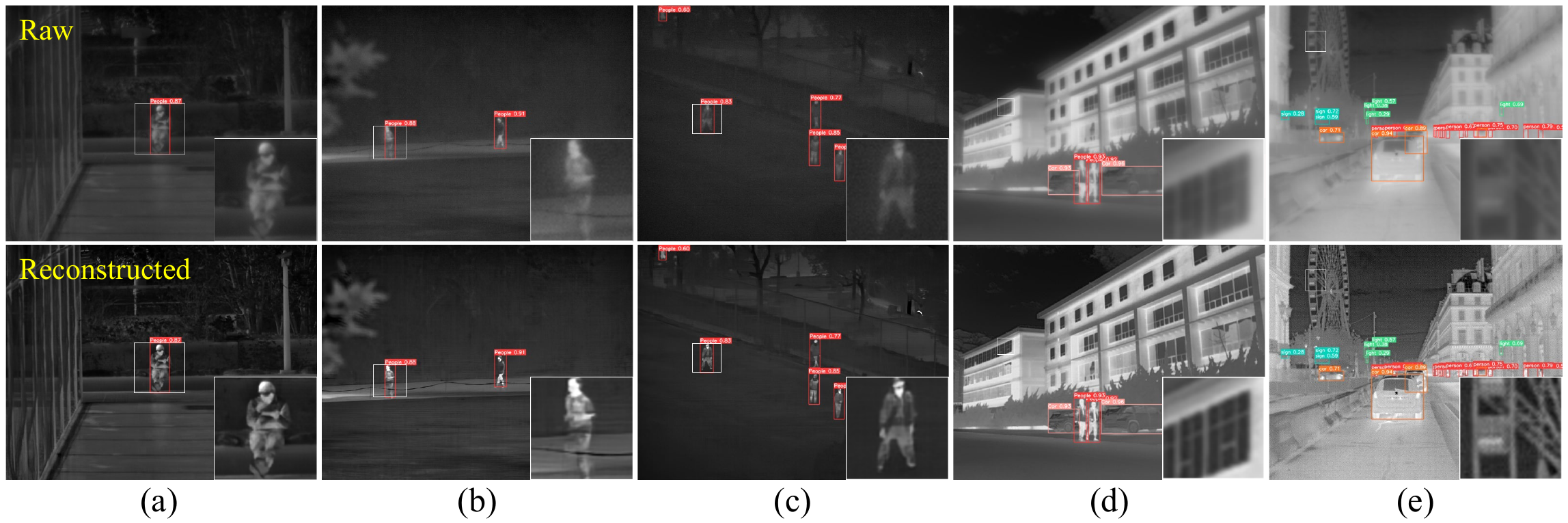}
		\caption{Detection results of the proposed PDI-Net with 50\% uniform structured pruning and INT8 precision quantization, visualized on the raw and reconstructed outputs: (a)--(c) captured by the deployed camera; (d) \textnormal{M}\textsuperscript{3}\textnormal{FD} dataset; (e) FLIR\_ADAS\_v2 benchmark.}
		\label{fig:edge-results}
	\end{figure*}
	
	Table~\ref{tab:comparison-flir} and Fig.~\ref{fig:comparison-flir} show results on the additional FLIR\_ADAS\_v2 dataset. The ranking is broadly consistent with \textnormal{M}\textsuperscript{3}\textnormal{FD}, although absolute metrics are lower, likely due to reduced image quality and annotation noise in FLIR\_ADAS\_v2. In the two rows, many targets are small, such as \textit{light} and \textit{sign}. ESOD, with its emphasis on small-object detection, produces more accurate boxes and higher confidence than IA-YOLO and DeepDenoising. The proposed PDI-Net slightly outperforms ESOD in the quantitative metrics and shows a clear advantage on \textit{light} and \textit{hydrant}. The same trend holds in cluttered or low-contrast scenes, indicating that physics-aware feature sharing preserves task-relevant structure even when global visual quality is weak. Performance differences between \textnormal{M}\textsuperscript{3}\textnormal{FD} and FLIR\_ADAS\_v2 likely reflect domain shift and class-label granularity mismatches; nevertheless, the stable ranking on both benchmarks supports the generalizability of the proposed PDI-Net. The zoomed insets exhibit trends highly consistent with those observed in the M\textsuperscript{3}FD dataset.

	\subsection{Scalability and compatibility}

	To verify the generality and scalability of the proposed PDI-Net, a YOLOv8s-based instantiation is evaluated under the same experimental settings. As shown in Table~\ref{tab:scalability-yolo}, the proposed PDI-Net achieves a competitive overall accuracy and markedly stronger performance at high IoU thresholds, confirming the compatibility and practicality of the framework beyond the YOLOv5s configuration. This is because the proposed PDI-Net is inherently detector-agnostic: the reconstruction encoder provides a multiscale feature pyramid, while the PALS-Bridge projects these physics-aware features into canonical tensor forms compatible with mainstream detectors. Consequently, replacing the detection head requires no modification to the joint objective $\mathcal{L}_{\text{total}}=hyp_{rec}\mathcal{L}_{\text{rec}}+hyp_{det}\mathcal{L}_{\text{det}}$, and only minor interface adjustments, such as channel alignment or anchor/free-anchor configurations. Conceptually, the PALS-Bridge enforces optical invariances associated with the PSF and sensor noise while preserving task-discriminative semantics. Therefore, any architecture that ingests a feature pyramid, including one-stage detectors (e.g., the YOLO family) and two-stage heads built on feature-pyramid networks, can be seamlessly integrated without altering the reconstruction pathway.

	\section{Outdoor experiments after edge deployment}
	To evaluate the on-device inference performance under real-world conditions, outdoor field experiments are conducted using a small UAV platform. The proposed PDI-Net is deployed on an RK3588 AI chip (Rockchip Electronics Co., Ltd.) integrated into a Matrice 350 RTK UAV (DJI-Innovations, China). To meet real-time constraints on resource-limited hardware, structured pruning with an average sparsity of 50\% is applied to eliminate redundancy and reduce computational overhead. The optimized model is then embedded into the single-lens infrared computational imaging system and mounted on the UAV for comprehensive validation. A supplementary video showcasing the online exhibition of the raw image output and real-time detection performance of the single-lens infrared computational imaging camera is provided via the \href{https://drive.google.com/file/d/1yo5lvQsTfszyAKVtQGpXztQPWVSXh5gm/view?usp=sharing}{online supplementary link}.
		
	\begin{table*}[!t]
		\centering
		\footnotesize
		\caption{Detection Results of the Proposed PDI-Net with 50\% Uniform Structured Pruning under Different Precision Quantization Settings}
		\label{tab:deployment-quantization}
		\begin{tabular*}{\textwidth}{@{\extracolsep{\fill}}lccccc}
			\toprule
			Quantization & Speed (ms) & Precision & Recall & mAP@0.5 & mAP@0.5:0.95 \\
			\midrule
			FP16 & 51.53 & \textbf{0.883} & \textbf{0.741} & \textbf{0.812} & \textbf{0.515} \\
			INT8 & \textbf{33.56} & 0.866 & 0.74 & 0.803 & 0.501 \\
			\bottomrule
		\end{tabular*}
	\end{table*}
			
	\begin{table*}[!b]
		\centering
		\footnotesize
		\caption{Specifications of the Single-Lens Infrared Computational Imaging Camera}
		\label{tab:camera-specs}
		\begin{tabular*}{0.8\textwidth}{@{\extracolsep{\fill}}lc}
			\toprule
			Parameter & Value \\
			\midrule
			System weight & 372 g \\
			Operating wavelength & 8--12~$\mu$m \\
			Focal length & 70 mm \\
			F-number & 1.0 \\
			Detector type & $\mathrm{VO}_x$ uncooled detector \\
			Array format & 640$\times$480 pixels \\
			Pixel size & 12~$\mu$m \\
			\bottomrule
		\end{tabular*}
	\end{table*}
				
	\begin{figure*}[!b]
		\centering
		\includegraphics[width=\textwidth]{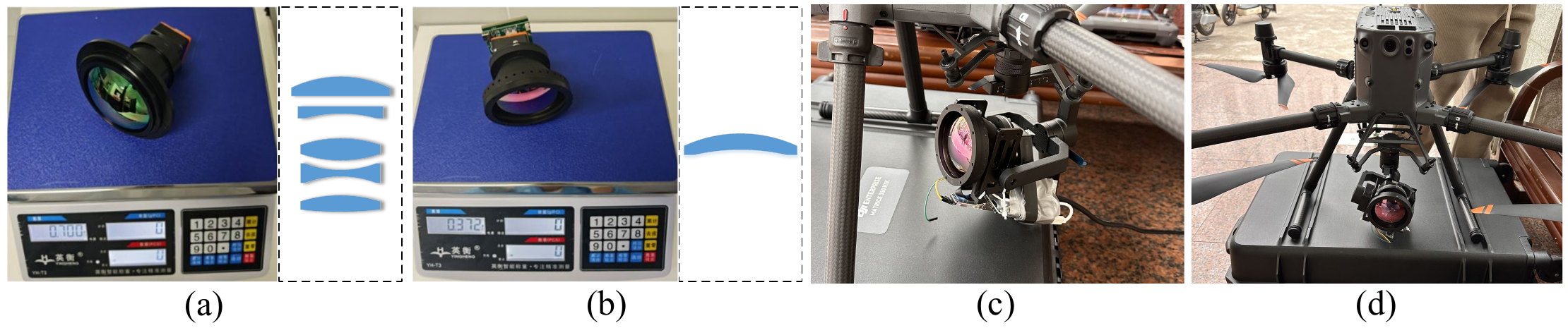}
		\caption{Imaging system comparison and UAV integration. (a) Traditional multi-lens infrared camera (700~g). (b) Proposed single-lens camera (372~g). (c), (d) UAV-mounted implementation of the proposed system.}
		\label{fig:uav-integration}
	\end{figure*}
	
	Table~\ref{tab:deployment-quantization} summarizes the performance of the proposed strategies deployed on the RK3588 platform. Both FP16 and INT8 quantization schemes were evaluated, where INT8 notably reduces memory consumption and accelerates inference while maintaining stable detection accuracy. Consequently, subsequent analyses are conducted using the INT8 on-chip quantization results. Fig.~\ref{fig:edge-results} presents imaging and detection outcomes, including cases with and without reconstructed images. Figs.~\ref{fig:edge-results}(a)--\ref{fig:edge-results}(c) depict representative scenes captured by the single-lens infrared camera, and the proposed method maintains stable detection performance across these examples. Figs.~\ref{fig:edge-results}(d) and \ref{fig:edge-results}(e) further demonstrate robust detection performance on benchmark datasets. Additionally, magnified insets reveal significant enhancements in image fidelity and structural clarity compared to the raw inputs. It should be noted that, in scenarios demanding high real-time performance and minimize latency, the proposed network omits the output of clear reconstructed images to minimize latency. Balancing human-perceptual image quality with low detection delay, such as through region-of-interest (ROI) reconstruction focused on target areas, represents a promising direction for future research and system optimization.

	\section{Single-lens infrared camera design and UAV integration}
	The key specifications of the proposed single-lens infrared computational imaging camera are summarized in Table~\ref{tab:camera-specs}. Compared with conventional multi-lens infrared imaging systems, the proposed system significantly reduces overall weight. As shown in Fig.~\ref{fig:uav-integration}(a) and \ref{fig:uav-integration}(b), the system weight is reduced from approximately 700~g to 372~g. This substantial reduction in payload is critical for UAV platforms, as it improves flight endurance and energy efficiency.

	This improvement is primarily attributed to the simplified optical architecture. Traditional infrared imaging systems typically rely on multi-lens elements to correct aberrations, resulting in increased system complexity and weight. In contrast, the proposed system adopts a single-lens design, where residual aberrations are compensated computationally by PDI-Net. This optical--computational co-design effectively reduces hardware complexity while maintaining imaging performance.

	The single-lens infrared computational imaging camera has been successfully integrated onto a UAV platform, as shown in Fig.~\ref{fig:uav-integration}(c) and \ref{fig:uav-integration}(d). The integration covers hardware interfacing, mechanical mounting, on-board computing, and communication. This system-level integration demonstrates the feasibility of the proposed PDI-Net for practical airborne infrared sensing applications.

	\section{Conclusions}
	This study presents a Physics-aware Dual-Integrated Network, named PDI-Net, which integrates infrared image reconstruction and object detection, while further integrating physical priors into the learning-based algorithm, thereby addressing the degradation introduced by lens simplification while satisfying low-latency requirements. The network consists of a U-Net-based reconstruction module and a YOLO-based detection module, with part of the encoder designated as a feature-sharing branch. In this way, full image reconstruction can be bypassed during inference, enabling efficient low-latency detection. These shared features are further routed through the proposed PALS-Bridge, which leverages physical priors from the imaging process to effectively bridge the gap between fidelity-oriented features and detection-oriented semantic representations. To support physics-informed training and evaluation, we develop a physics-informed simulation pipeline to synthesize optical degradation datasets based on \textnormal{M}\textsuperscript{3}\textnormal{FD} and FLIR\_ADAS\_v2. Experimental results demonstrate that, compared with the \textit{Rec+Det with pruning} strategy in the low-SNR setting, the proposed PDI-Net reduces inference latency by 84.06\% while improving mAP@0.5:0.95 by 5.07\%. Ablation studies further identify PALS-Bridge as the primary contributor to high-IoU gains while maintaining recall, and confirm that replacing the detection head requires only channel alignment and anchor reconfiguration. Finally, the proposed PDI-Net is deployed on an edge AI chip integrated with the proposed single-lens infrared camera, achieving an approximately 50\% reduction in system weight (from 700~g to 372~g) compared with traditional multi-lens configurations. In addition, the camera is integrated with a UAV platform for outdoor experiments, demonstrating the practicality of compact, real-time infrared computational imaging and target detection under resource-constrained conditions.

\section*{Acknowledgment}
This work was supported by the National Natural Science Foundation of China (Grant Nos. 62305250, 61925504, 62105243, and 62205248).

Supplementary material associated with this article is available online.

\bibliographystyle{IEEEtran}
\bibliography{Reference}

\end{document}